\documentclass[runningheads,a4paper]{llncs}

\usepackage{amssymb}
\usepackage{graphicx}
\usepackage{svg}
\usepackage{amsmath}
\usepackage{url}
\usepackage{multirow}
\urldef{\mailsa}\path|{s.eiffert,n.wallace,h.kong,navidp,salah}@acfr.usyd.edu.au|    
\newcommand{\keywords}[1]{\par\addvspace\baselineskip
\noindent\keywordname\enspace\ignorespaces#1}
\usepackage{float}

\begin{document}

\mainmatter  

\title{Experimental Evaluation of a Hierarchical Operating Framework for Ground Robots in Agriculture}

%
\author{Stuart Eiffert\thanks{Stuart Eiffert and Nathan Wallace contributed equally to this work; corresponding author: He Kong}, Nathan D. Wallace$^\star$, He Kong, Navid Pirmarzdashti, \linebreak and Salah Sukkarieh}

\authorrunning{Experimental Evaluation of a Hierarchical Operating Framework}
\titlerunning{Experimental Evaluation of a Hierarchical Operating Framework}


\institute{Australian Centre for Field Robotics, University of Sydney\\
\mailsa}

%
%

\maketitle

\begin{abstract}
For mobile robots to be effectively applied to real world unstructured environments---such as large scale farming---they require the ability to generate adaptive plans that account both for limited onboard resources, and the presence of dynamic changes, including nearby moving individuals. This work provides a real world empirical evaluation of our proposed hierarchical framework for long-term autonomy of field robots, conducted on University of Sydney's Swagbot agricultural robot platform. We demonstrate the ability of the framework to navigate an unstructured and dynamic environment in an effective manner, validating its use for long-term deployment in large scale farming, for tasks such as autonomous weeding in the presence of moving individuals.

\keywords{Field Robotics, Navigation, Dynamic Path Planning}
\end{abstract}

\section{Introduction}
Mobile robots are seeing increasing application in real world environments. However, for longer term deployments, the planning of the robot's operation must take its resource constraints into account, which, for applications such as large scale farming, may include energy and other materials necessary to perform tasks around the farm---such as herbicide for weed spraying. Consideration of these constraints is especially important due to the often broad scale of the areas the robot is expected to operate over. 

Additionally, to achieve robust autonomy, the robot must be able to adapt any mission level plan in response to the presence of both unknown and dynamic elements in these unstructured environments. Currently, autonomous field robots are limited to use in more structured environments, including row crops \cite{Bechar2017} or orchards \cite{Carpio2020}, where dynamic elements such as moving individuals are not a critical consideration. These approaches generally make use of reactive planners, which can adapt to unforeseen static obstacles. Planning in the presence of moving individuals, however, requires an understanding of how they might respond to the planned motion of the robot, in order to avoid the `frozen-robot problem' \cite{Trautman2010} which would otherwise occur in the presence of crowds and herds.

To address the above challenges, in our prior work \cite{Eiffert_CASE}, we have proposed a hierarchical framework for long-term and robust deployment of field ground robots in large-scale farming. In this paper, we present a real world evaluation of the former framework which integrates a local, interaction-aware dynamic path planner with a longer term resource aware mission planner to achieve long-term autonomy in unstructured and dynamic environments. Through trials conducted on the University of Sydney's Swagbot agricultural robot platform, we demonstrate that our framework is able to navigate an unstructured environment in an energy-efficient manner in the presence of moving individuals, enabling the completion of tasks such as weed spraying in farm environments.



\section{Technical Approach}
\label{section:planning}
\begin{figure}[t]
    \centering
	\includegraphics[width=12.2cm,height=4.4cm]{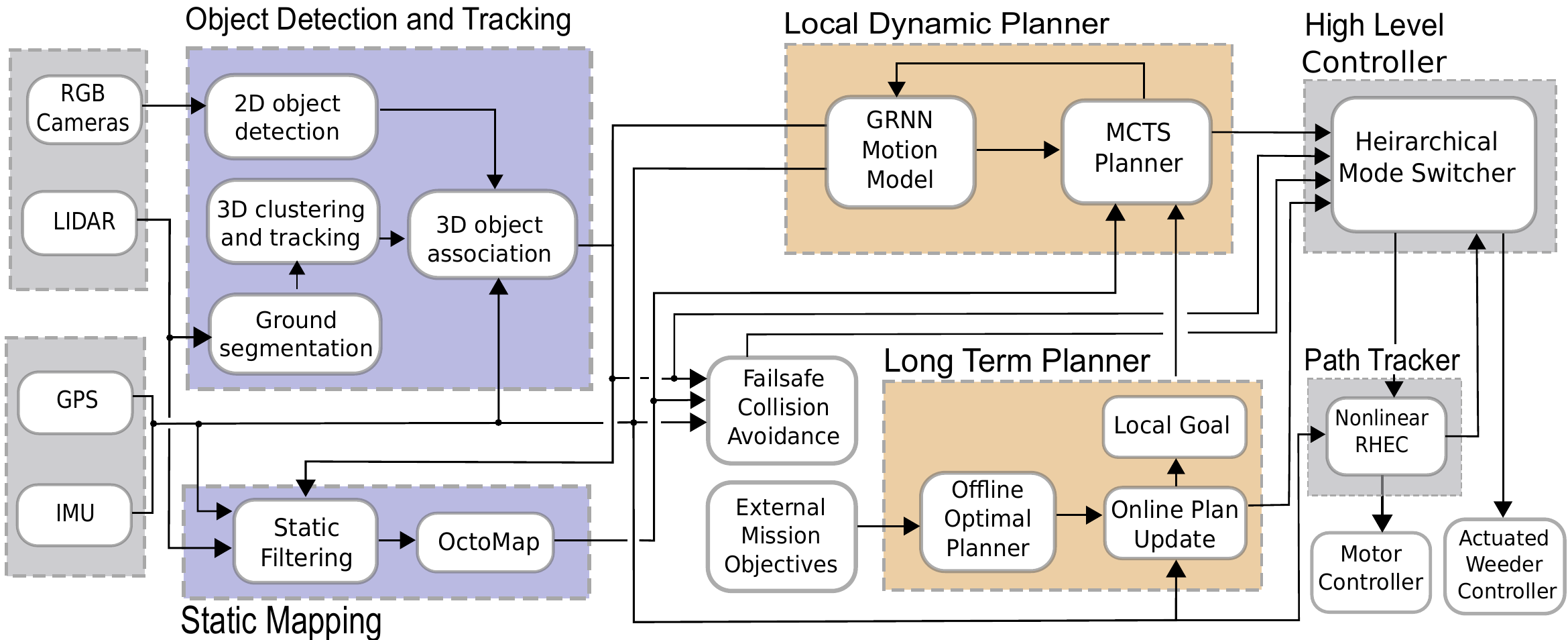}\vspace{-0.5cm}
	\setlength{\belowcaptionskip}{10pt}
	\caption{\textit{System overview of the hierarchical framework, illustrating the communication between each planner, the failsafe collision module, and the high level controller. Mission objectives are provided externally to the long-term planner.}}
	\label{system_overview}
\end{figure}

In our previous work \cite{Eiffert_CASE}, we developed a hierarchical framework for off-road, multi-objective robot navigation in the presence of both moving individuals and static obstacles. The performance of this framework was validated using simulated trials, in which moving agents were modelled using the ORCA reciprocal collision avoidance model of motion \cite{VanDenBerg2011}. In the present work, we further extend this hierarchical framework---shown in Fig. \ref{system_overview}---to include the operation of an actuated weeding arm, and evaluate its use in a real world agricultural environment, in the context of a spraying a set of pre-identified weeds in the presence of moving individuals.



\begin{figure}[t]
    \centering
	\includegraphics[width=10.0cm,height=4.5cm]{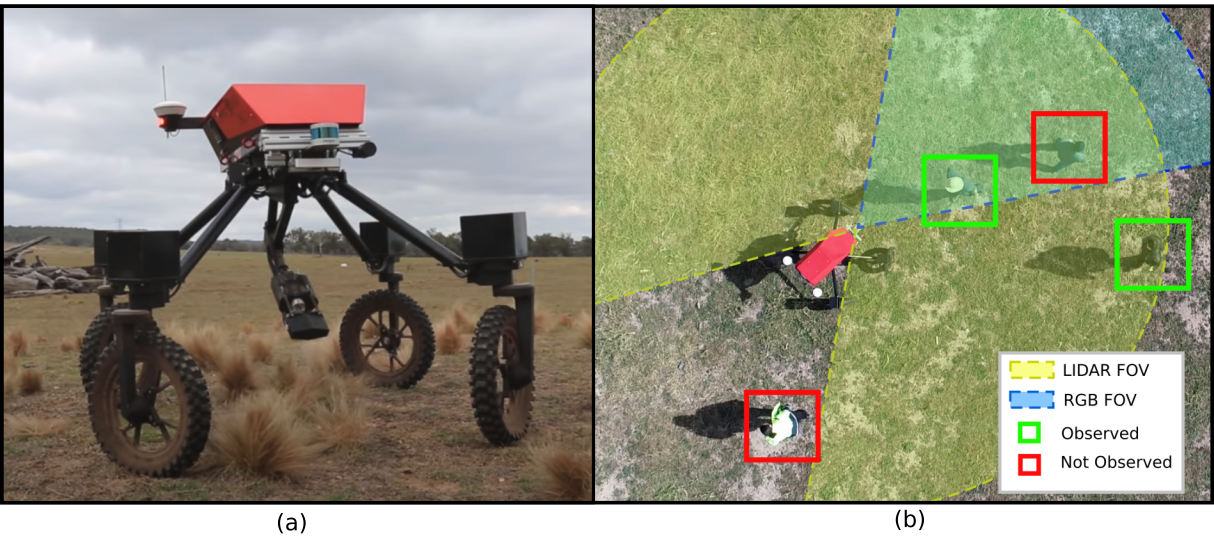}\vspace{-0.2cm}
	\setlength{\belowcaptionskip}{5pt}
	\caption{\textit{Swagbot Robotic Platform used for testing. (a) Photo of Swagbot with actuated weeder extended. (b) Top-down illustration of sensor FOV, and the blind spot arising due to the front-mounted sensor installation.}}
	\label{robotic_platform}
\end{figure}

The hierarchical framework is composed of a few key components: an object detection and tracking module, a long-term planner, a local dynamic planner, a high level switching controller, and a path tracker. 
The multimodal perception solution used in this work utilises both LIDAR and RGB for object detection, as per our simulated work in \cite{Eiffert_CASE}. However, due to limited sensor field of view, as shown in Fig. \ref{robotic_platform}, detection is not possible directly behind the robot. The tracking module used has been updated to be aware of the sensor field of view, ensuring that updating of tracked object confidence based on missed observations is performed at a slower rate when it enters the blind spot, in order to maintain object permanence.

The long-term planner is based on \cite{Wallace_ICRA}--\cite{Wallace_WROCO}, in which an energy efficient path is generated by searching over a Probabilistic Roadmap of the environment to find the lowest cost paths between all the goal waypoints---determined using a learnt energy cost of motion metric---and by then solving an asymmetric Travelling Salesman problem over this emergent graph to yield the optimal plan (alternatively, the long-term plan can also be generated using our recent work in \cite{OPR} on the orienteering problem with replenishment, accounting for when, where, and for how long to recharge resources). This plan takes into consideration the topography of the area, the slip conditions of the terrain, and any previously known non-traversable areas. The local dynamic planner used in this work is adapted from our prior work in \cite{Eiffert_ICRA,Eiffert_ACRA}, using generative Recurrent Neural Networks (RNNs) and Monte Carlo Tree Search (MCTS). A high level controller switches between planning modes based on proximity to detected moving individuals, static obstacles, and waypoints.
\vspace{-0.3cm}

\section{Experimental Evaluation}
\begin{figure}[t]
    \centering
    \includegraphics[width=10.0cm,height=10.0cm]{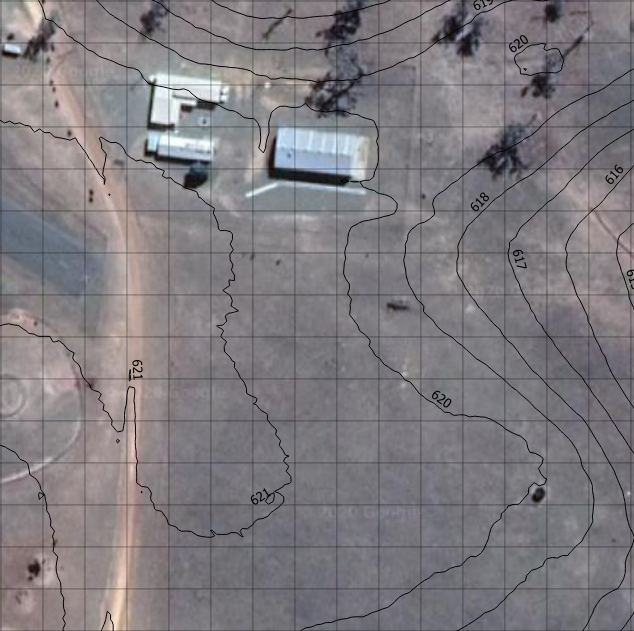}
    \setlength{\belowcaptionskip}{5pt}
    \caption{\textit{Satellite imagery of the Arthursleigh Farm test site, where the experiments were conducted. Terrain topography indicated by the contour lines, and spatial extent indicated by grid lines spaced at 20m intervals.}}
    \label{trial_site}
\end{figure}

Real world empirical evaluation of our approach has been conducted on the University of Sydney's Swagbot agricultural robot platform; a four-wheeled independently driven and steered electric ground vehicle capable of omnidirectional motion, which was designed for use in environments with uneven terrain. Testing was performed at the University's Arthursleigh Farm---pictured in Fig. \ref{trial_site}---and involved the continuous navigation between updated sets of externally provided mission waypoints across an unstructured field 2 ha in size. At each waypoint the robot was required to accurately spray a pre-located weed, operating in the presence of both moving individuals and unknown obstacles. Each iteration of the trial began at a home waypoint, where a set of 5-8 objective waypoints were supplied to the robot. 

\begin{figure}[t]
    \centering
	\includegraphics[width=11.0cm,height=10.0cm]{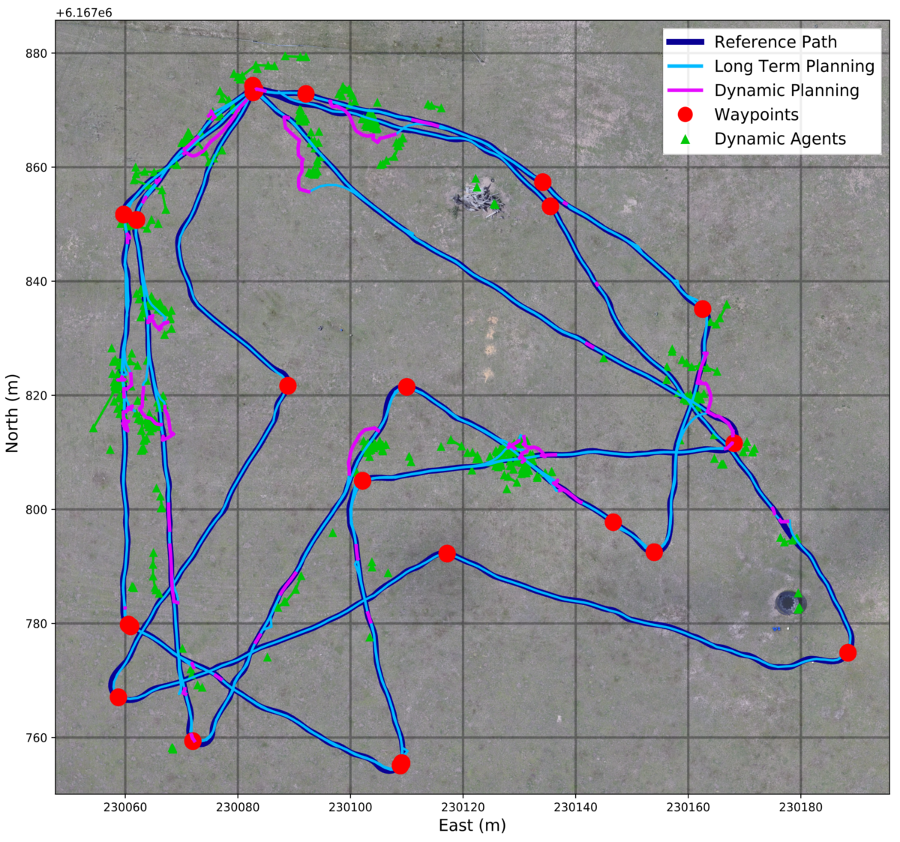}
	\setlength{\belowcaptionskip}{5pt}
	\caption{\textit{Complete course, showing repeated route of the robot during the trial, all waypoints, and detected agents. The reference path (dark blue) is overlayed with the actual taken path, differentiating between when the long-term planner module (light blue) and dynamic planning module (purple) were each in use.}}
	\label{trial_map}
\end{figure}

As per Section \ref{section:planning}, an optimal energy-efficient path was then computed and used as the reference path for online local planning during navigation to each objective. 
Upon returning to base, a new set of waypoints were supplied and the trial repeated. A total of 3 sets of waypoints were reused, with testing continuing until the robot exhausted its energy resources. An overview map of the testing area showing the repeated route of the robot is illustrated in Fig. \ref{trial_map}.
Total time of the trial was 2hr44mins, covering a distance of 5.49km, including 37 distinct interactions with groups of moving agents. For the purposes of this trial, people---rather than livestock---were used as agents, both for safety, and due to the availability of a response prediction model of person-robot interactions from our prior work. Performance was determined using metrics of: (1) number of near-collisions with moving individuals; (2) average speed between waypoints; (3) deviation from the reference path.

\section{Results}
 \begin{figure}[!ht]
    \centering
	\includegraphics[width=12cm,height=4.5cm]{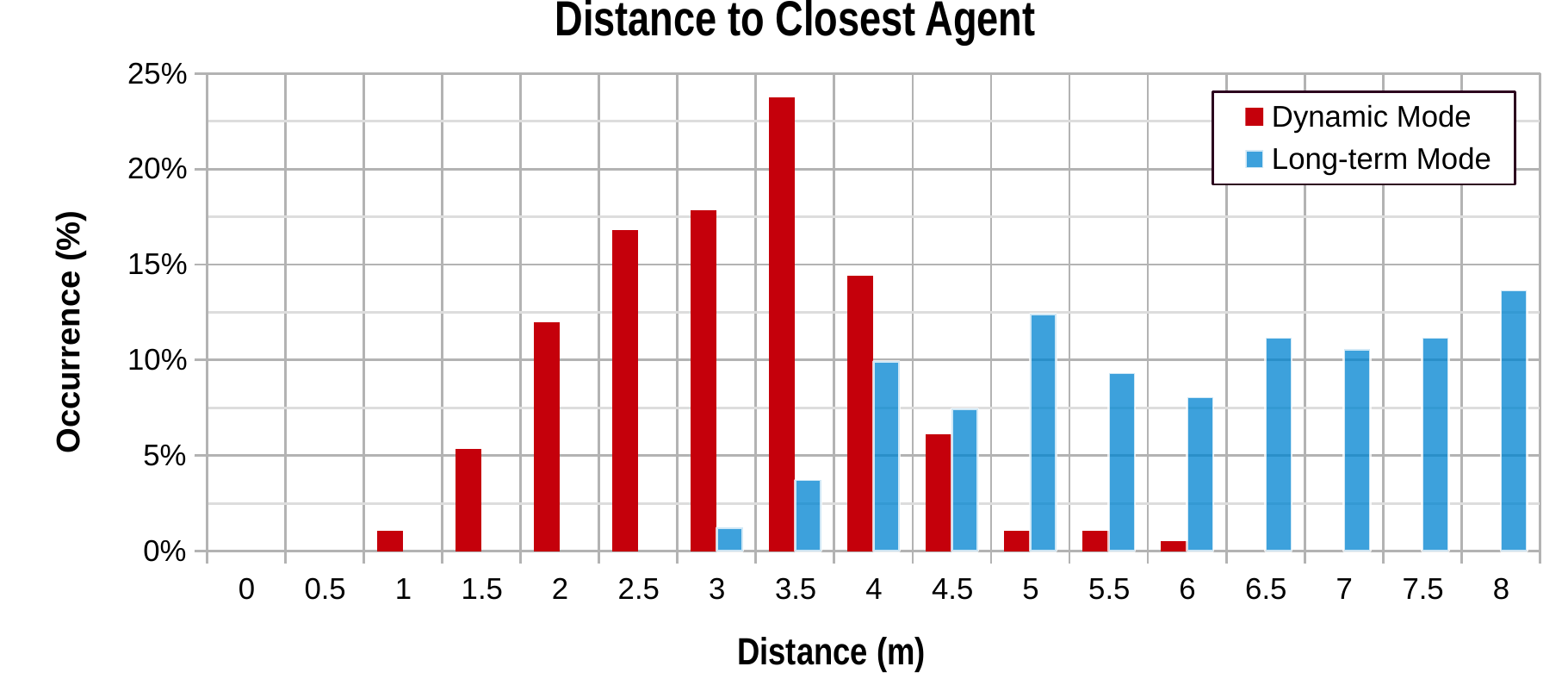}
	\setlength{\belowcaptionskip}{12pt}
	\caption{\textit{Histogram of distances to closest agent by planner mode, shown as a percentage of occurrence of times when an agent was within 8m. Dynamic planning mode is able to successfully take over from the long-term planner when in the presence of other agents and avoid collisions. Mode handover is dependent both on proximity of agents to the robot and to the prior planned path, resulting in the overlap shown.}}
	\label{distances}
	\includegraphics[width=12cm,height=5cm]{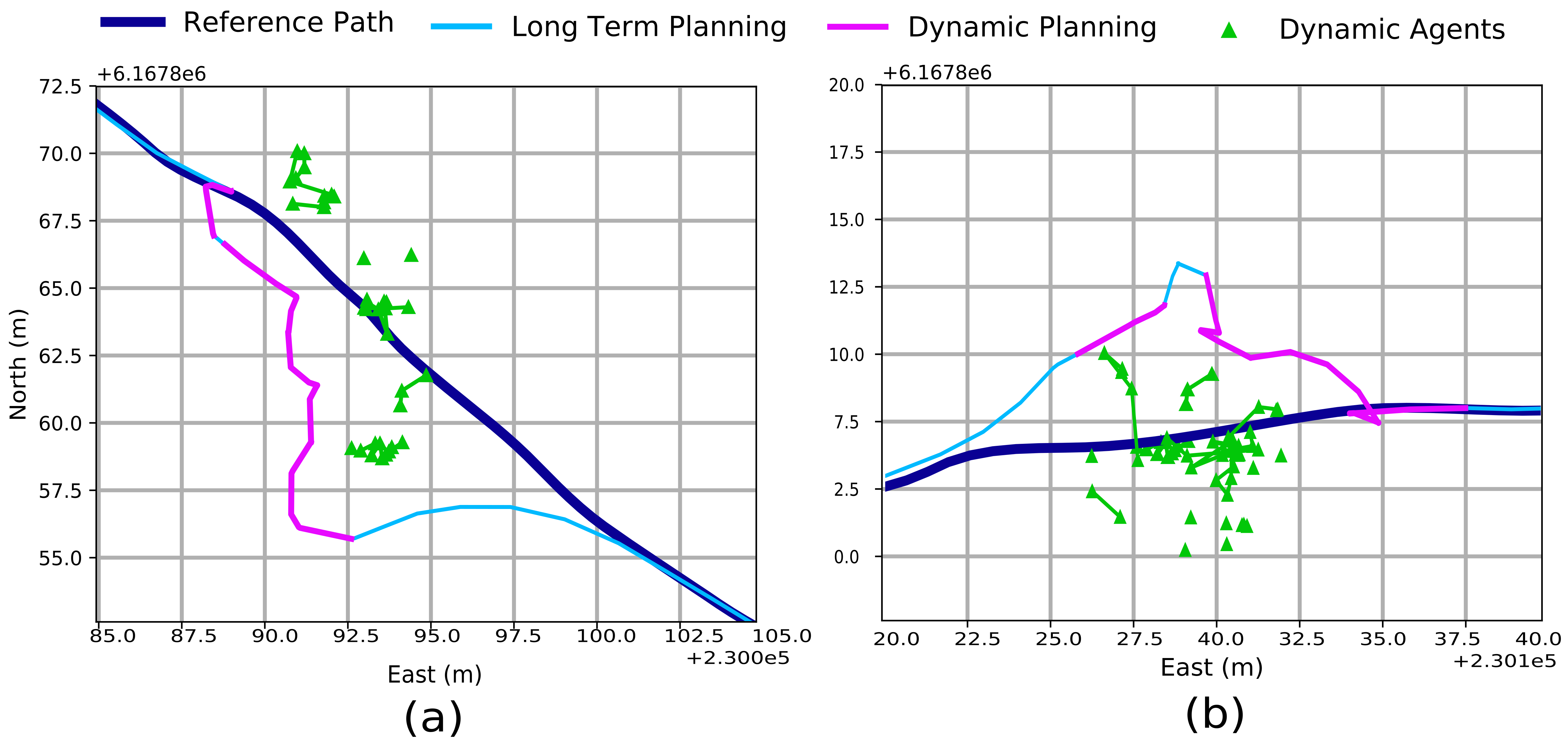}
	\setlength{\belowcaptionskip}{10pt}
	\caption{\textit{Navigation around moving individuals, highlighting when the dynamic planning module was in use (purple). (a) the robot initially attempts to pass between agents, but instead takes wider route as the agents continue their path down. (b) less desirable behaviour, where tracking of nearby agents failed when the robot was facing away. Both examples show the robot returning to the reference path when clear of agents. }}
	\label{dynamic_examples}
\end{figure}

\subsection{Dynamic Collision Avoidance}
Throughout the trial 9.42\% of time was spent in dynamic planning mode (a total of 15.5 minutes). Switching between dynamic and long-term planning modes is handled by the high level controller as shown in Fig. \ref{system_overview}, dependent on proximity of detected agents both to the robot and the prior output of the path tracker module. As shown in Fig. \ref{distances}, this allows the robot to continue in long-term mode even when in close proximity to nearby agents if the current planned path will not result in a collision. Fig. \ref{distances} also demonstrates the ability of the dynamic planning mode to successfully avoid collisions and navigate around dynamic agents. The robot was able to maintain an average distance of 3.36m from all agents whilst in dynamic mode. Fig. \ref{dynamic_examples} (a) shows an example of this desired behaviour, whilst example (b) shows the single occasion in which an agent came within the fail-safe distance of 1.5m, reaching a minimum distance of 1.30m before the robot stopped.
During this interaction the robot turned on the spot without realising that a person was present in its `blind spot'. This occurred both as a result of limited sensor field of view and inconsistent implementations of motor controller between prior simulated work in \cite{Eiffert_CASE} and this experiment. In \cite{Eiffert_CASE}, planning was done using a non-holonomic simulated robot configured for Ackermann motion, limiting motion to forward direction only. However, the real world robot is capable of omnidirectional motion, and while the planner constrained the robot to drive forwards, preventing it from travelling into its blind spot, it could nevertheless still turn on the spot whilst tracking both the dynamic and long-term paths.

 \begin{figure}[t]
    \centering
	\includegraphics[width=12cm,height=6cm]{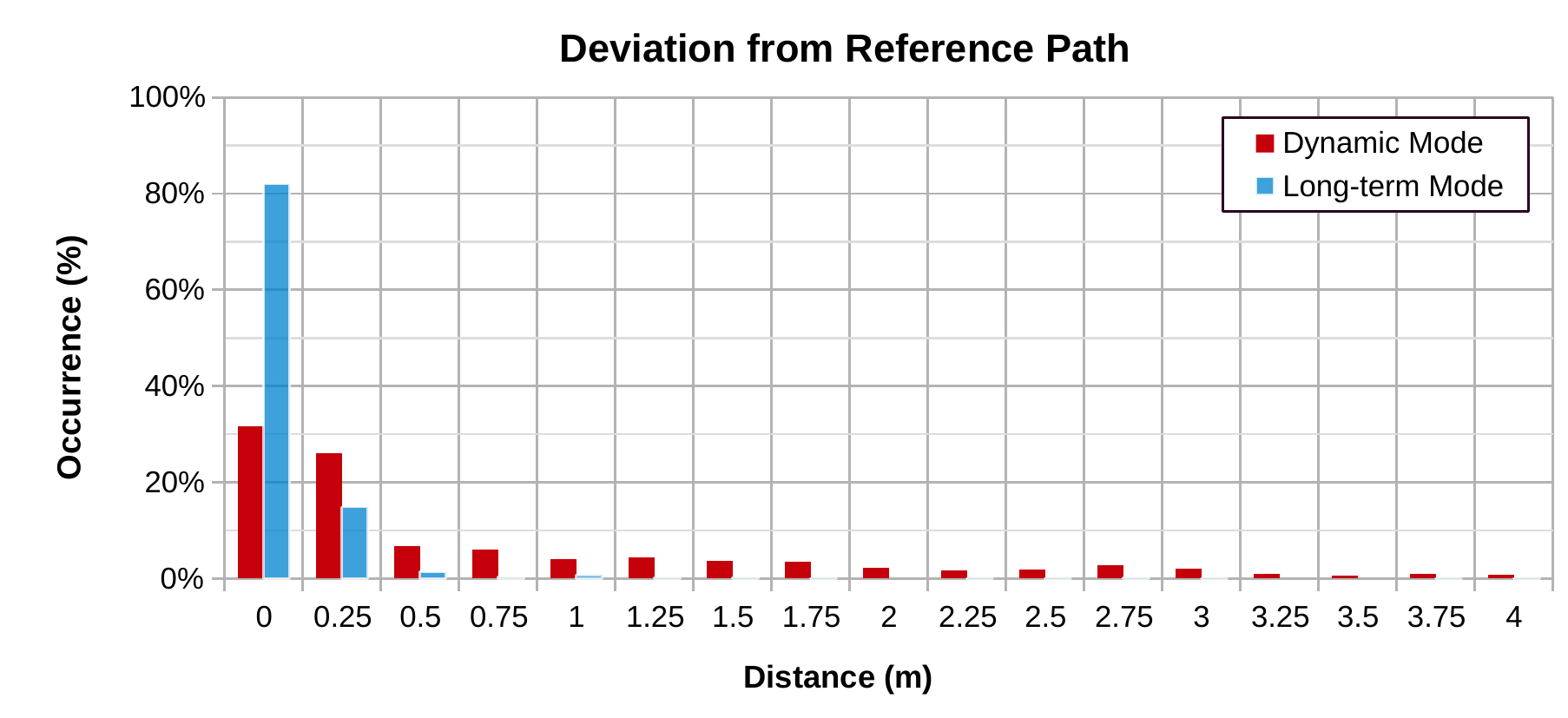}
    \setlength{\belowcaptionskip}{15pt}
	\caption{\textit{Distances from the offline computed energy optimal reference path throughout the trial, by planner mode. Both the long-term and dynamic mode tend to remain close to the reference, however the dynamic planner can deviate significantly in order to navigate around nearby moving individuals, as shown by the protracted tail.}}
	\label{deviations}
\end{figure}

\begin{figure}[!ht]
    \centering
	\includegraphics[width=12cm,height=12cm]{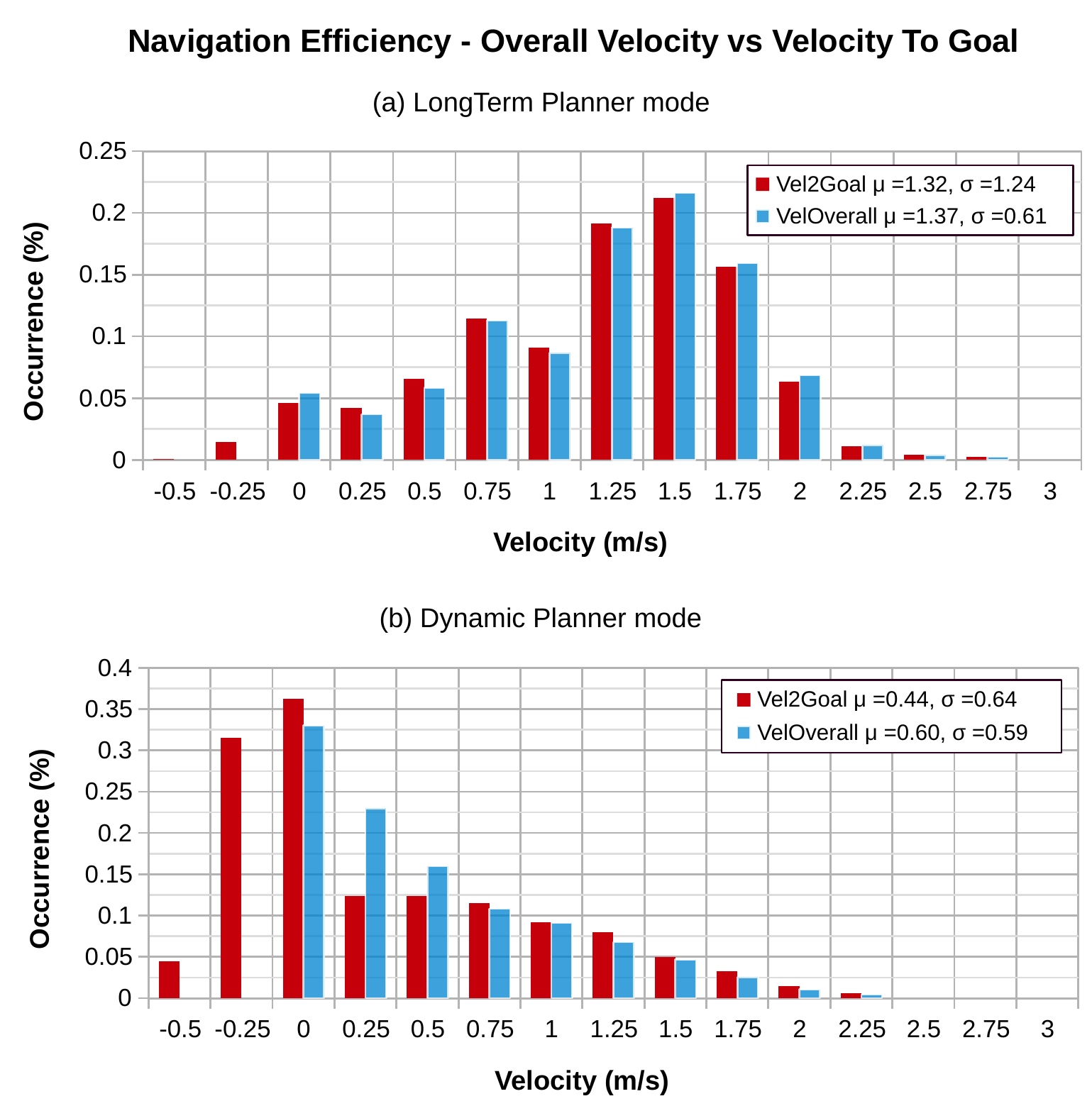}
		\setlength{\belowcaptionskip}{15pt}
	\caption{\textit{Comparison of overall velocity (VelOverall) and velocity along the reference path (Vel2Goal) when in long-term planning mode (a) versus dynamic planning (b) illustrating the impact of nearby moving individuals on the robot's ability to follow the reference path. The average VelOverall in dynamic mode is less than half of that achieved in long-term mode. Additionally, the ratio of Vel2Goal to VelOverall is significantly less in dynamic mode, with over 35\% of time spent travelling away from the optimal direction.}}
	\label{speeds}
\end{figure}

\subsection{Navigation Efficiency}
\label{section:efficiency}
Fig. \ref{deviations} illustrates how significant deviations from the prior optimal path can occur whilst navigating through groups of individuals, although for the most part the dynamic planner tends to remain close to the reference path when possible. This is desirable since keeping close to the long-term reference path will retain the energy efficiency merits and minimise resource usage.
Similarly, a comparison of the velocity of the robot along the optimal path, compared to its overall velocity, allows an analysis of how much energy is required navigating around dynamic obstacles. Fig. \ref{speeds} highlights the impact that the presence of moving individuals has on the efficiency of the robot's navigation. As shown in (a), when in long-term planning mode, almost all of the robot's velocity is directed along the optimal path, with average velocity towards goal of 1.32 m/s nearly matching average overall velocity of 1.37 m/s. However when dynamically navigating around moving agents the robot achieves an overall velocity of only 0.60 m/s. Additionally, a significant amount of energy is directed away from the optimal path, as shown in (b). Over 35\% of motion is directed away from the local goal resulting in an average velocity towards goal of 0.44m/s, which is 33.3\% of that achieved in long-term mode.

\section{Discussion}
This work has shown that the proposed system allows for safe operation of a mobile robot around moving individuals during the completion of tasks such as weeding an agricultural field. Overall, the presented results have successfully demonstrated the implementation of our hierarchical framework in the real world. These results validate the hierarchical framework for use in large scale farming, allowing the safe and efficient operation in dynamic and unstructured environments.

However, a number of outstanding challenges remain in order to extend the proposed system for long-term autonomy. Improved perception around moving individuals is required, including the ability to better detect and track agents on a robotic platform with a limited field of view, and the proper integration of the planner's dynamics and the real world platform dynamics. Additionally, real applications of long-term autonomy require inclusion of a recharging station, rather than simply designating a base waypoint as such as was done in this work.  Whilst other works \cite{varandas2019} have shown how this can be applied to mobile robots in agriculture, it will need to be included in the proposed hierarchical framework in future.
Based on the results discussed in Section \ref{section:efficiency}, an understanding of how dynamic adjustments to the offline computed optimal path influence energy usage and navigation efficiency should also be considered. This suggests the need for consideration of crowd or herd density during extended operation, to ensure mobile robots are able to effectively reach a recharging station when required.

\end{document}